# VISUAL LOCALISATION OF MOBILE DEVICES IN AN INDOOR ENVIRONMENT UNDER NETWORK DELAY CONDITIONS

Alberto Alonso Fernández[1], Omar Álvarez Fres[1], Ignacio González Alonso[1] and Huosheng Hu[2]

[1]Computer Science Department, Oviedo University, Oviedo, Spain
`alonsoalberto@uniovi.es, alvarezomar@uniovi.es,`
`gonzalezaloignacio@uniovi.es`
[2] School of Computer Science & Electronic Engineering, University of Essex, Colchester CO4 3SQ, United Kingdom
`hhu@essex.ac.uk`

## ABSTRACT

*Current progresses in home automation and service robotic environment have highlighted the need to develop interoperability mechanisms that allow a standard communication between the two systems. During the development of the DHCompliant protocol, the problem of locating mobile devices in an indoor environment has been investigated. The communication of the device with the location service has been carried out to study the time delay that web services offer in front of the sockets. The importance of obtaining data from real-time location systems portends that a basic tool for interoperability, such as web services, can be ineffective in this scenario because of the delays added in the invocation of services. This paper is focused on introducing a web service to resolve a coordinates request without any significant delay in comparison with the sockets.*

## KEYWORDS

*Location, Interoperability, WebServices, MoCap, Sniffer, Chokepoint.*

## 1. INTRODUCTION

DHCompliant project arises from the need to promote the implementation and integration of home automation and robotic technologies at home. The objective of this project is to study, to specify and to develop a communication standard between these systems. DHCompliant protocol is divided into a number of subsystems that can meet existing needs in a home automation environment. DHC-Localization is one of the subsystems within this protocol.

DHC-Localization [1] is able to provide the necessary data for a mobile robot in order to reach a destination. In other words, it provides the robot location information needed to establish a route between two points, the current robot location and the point where the robot must perform a moving. In order to perform this task, it will use an adjacent system which continuously calculates the position of the scenario's moving parts.

The communication among DHC central system, robot adapters and the coordinates measurement adjacent system can be implemented in multiple ways. Sockets are the most traditional connection mechanism [2], but don't pose an optimal solution in terms of interoperability between heterogeneous systems, because the socket implementation must have an explicit selection of the protocol, the destination IP address and the destination port. It is also necessary to know the type of data that will be returned, which means that both sides must have mutual understanding [3].

On the other hand, web services are the interoperability tool that will be used as entry to the Internet of Things concept [4] on which the future digital homes will be based. Web services allow knowing the information needed to invoke the service and the returned results thanks to the WSDL description file [5]. The delay added in the invocation of the web services can be a problem in order to obtain real-time data from the coordinate measurement adjacent system [6].

This article presents the coordinate measurement systems being deployed and the connection methods. Some experiments are conducted in order to identify the advantages and disadvantages of DHC-Localization and the dependence of the coordinate measurement system. It's important to perform a preliminary study in order to assess the impact and the dependence of the coordinate measurement system and the DHC-Localization subsystem before the real implementation of DHCompliant is made.

## 2. STATE OF THE ART

In this section it will be analyzed the coordinates measurement systems that can be used in conjunction with the DHC-Localization subsystem. These measurement systems are heterogeneous and use different technologies that should provide sufficient accuracy to locate subjects in indoor environments.

Due to the fact that the DHC-Localization subsystem implementation must be independent from the coordinate measurement adjacent system, different connections will be contemplated between them.

### 2.1. Coordinate Measurement Systems

In order to calculate the position of mobile devices in a home environment, it's possible to use different technologies for instance the GPS, however this technology doesn't work inside buildings [7], that's why other possibilities such as motion capture, artificial vision or the use of locator signals, are considered [8].

### 2.1.1 Real-time Location System (RTLS)

Real-time Location Systems [9] are fully automated systems that can constantly monitor the locations of devices and robots. At the beginning, the RTLS systems used the unlicensed spectrums (408, 433, 900 MHz). After the emergence of the IEEE 802.11, the communication between the tag-to-access point air interface and the 802.11, an access point (AP) is facilitated. Three major technological concepts enable RTLS to determine a client location:

*Nearest Access Point*

It is used in environments to detect the presence of an asset within a specific area. A single Access Point or Location Receiver can detect tags and WiFi devices.

*Time Difference on Arrival (TDoA)*

It relies on the time a packet needs to be sent from a transmitting device towards a receiving one. When the device to be located sends out a time stamped signal using WiFi, this signal is received by the access points. Once the time difference is known, it's possible to calculate the distance between the access point and the device. The location of a device can be estimated in a system with at least three APs.

*Received Signal Strength Indicator (RSSI)*

It measures the signal strength of the access points and sends the values to a server. The server is responsible for calculating the coordinates of the device once it has three measurements.

While TDoA is vulnerable to signal reflection caused by various obstacles in an indoor environment, the RSSI technology is optimal for indoor environments with walls [10].

Hara indicates in [11] that another advantage of the RSSI technology over the TDoA is that the TDoA requires the installation of an AP able to calculate the time difference measurement which implies a hardware spending, while the RSSI solution is based on an independent software and hardware solution [12].

The big RTLS vendors at the moment are AeroScout, Awarepoint, Ekahau, GE (Agility), InnerWireless (PanGo), Radianse, Sonitor and Versus [13]. From all these systems, Ekahau RTLS is the industry only WiFi based Real-time Location System that operates over any standard enterprise WiFi networks without the need to add proprietary sensors, chokepoints or additional WiFi radios. It is capable of pinpointing floor, room and sub-room levels with accuracy for asset, user and devices tracking, without impacting on the performance or use of the WiFi network.

### 2.1.2 Computer Vision (OpenCV)

Computer Vision can transform data from a still or a video camera into a grid of digital numbers so that useful decisions can be made about real physical objects and scenes based on sensed images [14]. Unlike the human vision that is processed and interpreted by the brain through a division of visual signals through a cross partnership resulting of life experience, the computer vision only generates a grid of numbers.

OpenCV is a free library for machine vision and image processing written in C, it runs under Linux, Windows and Mac OS X. It was initially developed as an Intel Research initiative to advance on CPU-intensive applications. With the expansion and opening of projects like real-time ray tracing and 3D display walls, OpenCV was conceived as a way to make computer vision infrastructure universally available [15]. The open source license for OpenCV allows building a commercial product using all or part of OpenCV library.

The library covers several areas such as 2D and 3D toolkits, egomotion estimation, facial recognition system, gesture recognition, Human-Computer Interface (HCI), mobile robotics, motion understanding, object identification, segmentation and recognition, stereo vision (depth perception from 2 cameras), Structure from Motion (SFM) or motion tracking.

Using OpenCV to calculate coordinates is based on the recognition by a camera of graphic patterns located on mobile devices. The process requires the system calibration to convert the pixels of the captured image in a length measure. These length measurements will define the mobile devices coordinates and orientation.

### 2.1.3 Motion Capture (Vicon System)

The motion capture or motion tracking is the process in which a movement is recorded and translated into a digital model. Its application area includes the military, entertainment, sports, medical applications, and the validation of computer vision and robotics [16].

Earliest motion capture systems used mechanical armatures to measure joint angles. Nowadays, there are two major techniques used to capture the movement of a device: Optical Systems and Non-optical Systems. Optical tracking systems use special visual markers on the performer and a number of special cameras are used to determine the 3D location of the markers. The main types of markers are:

*Passive Markers*

Usually the markers are passive objects such as retroreflective spheres. These spheres are monitored by a series of high speed cameras (between 4-32), monochrome devices tuned to sense a specific colour of light. These cameras are surrounding the object or device to be captured. Each camera is running at 30-1000 frames per second.

*Active Markers*

In this case the markers are LED lights. Only one marker is illuminated at a time, helping the system to keep track of the markers. The marker is powered to generate their own light, that's why this marker system produces an increase in the distance and volume for motion capture. Optical systems require multiple cameras to see a marker in order to triangulate its position, and may drop markers due to occlusion.

From motion capture mechanisms based on non-optical systems, it is necessary to emphasize the magnetic motion capture technology. This technology uses transmitters that establish magnetic fields within a space; it also uses sensors that can determine the position and orientation of individuals within the space, based on these fields. Each marker has its own sensor data channel.

Unlike a magnetic system, an optical system must determine the correspondence of markers between frames. This is done in post-processing software which is based on continuity of positions. Hardware solutions use active markers such as miniature LEDs to disambiguate markers [17].

The Vicon System [18] is a commercial solution for motion capture, and can be used such as a tracking system for recording robot positioning data [16]. It is based on the simultaneous recording by 8 CMOS video cameras of small reflective balls (markers) which are attached to the mobile robot or to the mobile device. This optical 3D system has less than 10 ms of latency, a positional accuracy of 0.1 mm and an angular accuracy of 0.15°. It can track up to 50 and 150 markers. The coordinates of the robot are obtained from these markers.

## 2.2. Connections to the Coordinate Measurement System

Each coordinate measuring system shall have a different connection method. The objective is to minimize the time consumed by the connection and achieve the most innovative alternative. In many cases this connection can be a chokepoint in our DHC-Localization system. The network programming techniques in a computer networking course addresses sockets, remote procedure call, and object-oriented remote procedure call. Web services appear to meet the need to standardize communication between different platforms and programming languages [19].

The main advantage offered by web services is the platform and programming language independent, which is a W3C standard, web services also add functionality to web sites, and use the HTTP protocol for transport and standardized elements for each of its components (SOAP, UDDI, WSDL, XML) [20]. These advantages distinguish web services from sockets making possible that WS can have an interface with the methods offered by the web service, however the sockets are reduced to just a transmission of packets. The use of HTTP also allows Web Services to cross the security restrictions of network routers.

On the other hand the biggest problem of the web services is the compliance or not of a quality of service requirements. It is especially remarkable the response time QoS property [21]. The response time is the time a service takes to respond to various types of requests. Response time is a function of load intensity, which can be measured in terms of arrival rates (such as requests per second) or number of concurrent requests. QoS takes into account not only the average response time, but also the percentile of the response time [22]. This is why it must be meticulously studied the impact of web services on the localization system quality. As a real-time system, any factor that adds a time delay into the DHC-Localization can be considered as a quality of service detriment.

For example, if our robot (which is a mobile device) makes a coordinates request and the response is not received in time, there will be a big difference between the actual position of the robot and the coordinates in which it believes to be.

## 3. METHODOLOGY

### 3.1. Design of the Experiment

A Vicon System with eight CMOS cameras is deployed in a robotic arena at Essex University, those cameras are positioned around the robotic arena at a height of 4.80 m. The Vicon System has been used to perform this research experiment, as shown in Figure 1. The robot has a set of five markers placed on the robot body to track its position. The robot position is determined from these markers in a form of (x, y, z, ψ) which corresponds to 3D positions and yaw angle of the robot.

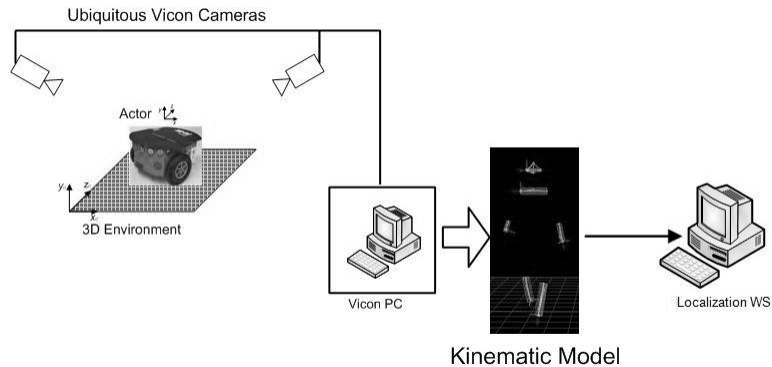

Figure 1. System configuration

Each robot has a TCP socket client for the Vicon System to make a connection and obtain the location data. This type of connection is a restriction added by the Vicon motion capture system. In order to study the time added for a web service in the process of obtaining coordinates, it has been built a location web service where each robot is a web service client and consume the web service it (see Figure 2).

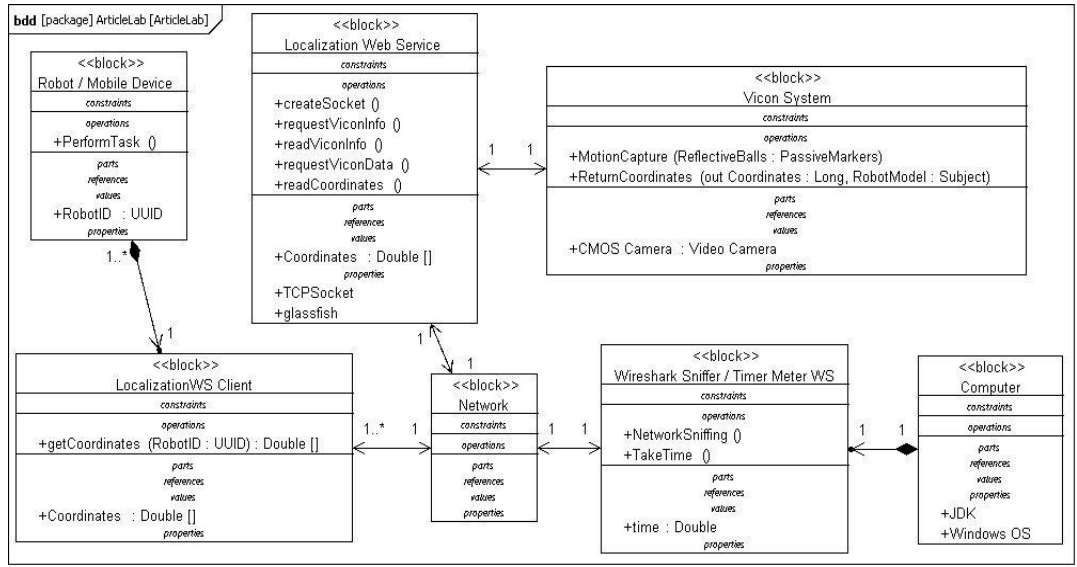

Figure 2. Capture Arena SysML Block Diagram

A time-measuring tool (e.g. Wireshark network sniffer [23] or a web service created for the purpose) is deployed for recording the data from two different types of connection mechanisms by using the motion capture system, those mechanisms are a Socket and a Web Service (with an underlying TCP socket). 100, 500 and 1000 iterations are made for each type of connection, the time measurements are taken in order to compare the two different connection mechanisms (see Figure 3).

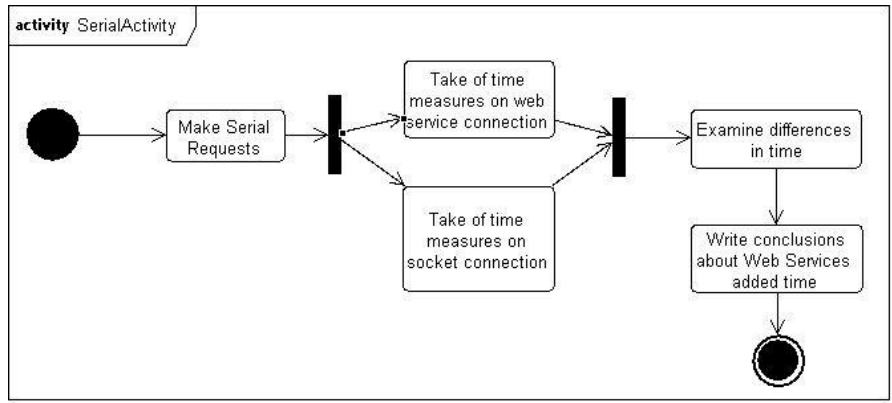

Figure 3. Serial Requests. SysML Activity Diagram

The configuration of the equipments used for testing is shown in Table 1.

Table 1. Equipment Configuration

|  | **Web Service Client** | **Localization Web Service** | **Vicon Client** |
|---|---|---|---|
| **CPU** | Intel Pentium 4 3.00 GHz | Intel Xeon 5150 2.66 GHz | Intel Xeon 5150 2.66 GHz |
| **RAM Memory** | 2.00 Gb | 2.00 Gb | 2.00 Gb |
| **Disk** | 75 Gb | 500 Gb | 500 Gb |
| **O.S.** | Windows XP SP3 | Windows XP SP3 | Windows XP SP3 |
| **Development Kit** | JDK 6 Update 21 | JDK 6 Update 21 | JDK 6 Update 21, Nexus 1.3 |
| **Application Server** | Glassfish 3.01 | Glassfish 3.01 | Glassfish 3.01 |
| **IDE** | NetBeans IDE 6.9 | NetBeans IDE 6.9 | NetBeans IDE 6.9 |

It is predictable that the socket needs less time than the web service. Moreover, the experiment aims to test the ability of the coordinate measurement system to execute requests in parallel. For this reason, a capture of times will be realized and the system behaviour will be analyzed on different moments of the experiment, the analysis will take place when having received 5, 10, 20, 50, 100, 250, 500, 1000 coordinates requests simultaneously (see Figure 4). The results of these experiments will allow considering the level of dependence existent between the DHC-Localization and the coordinate measuring system underneath.

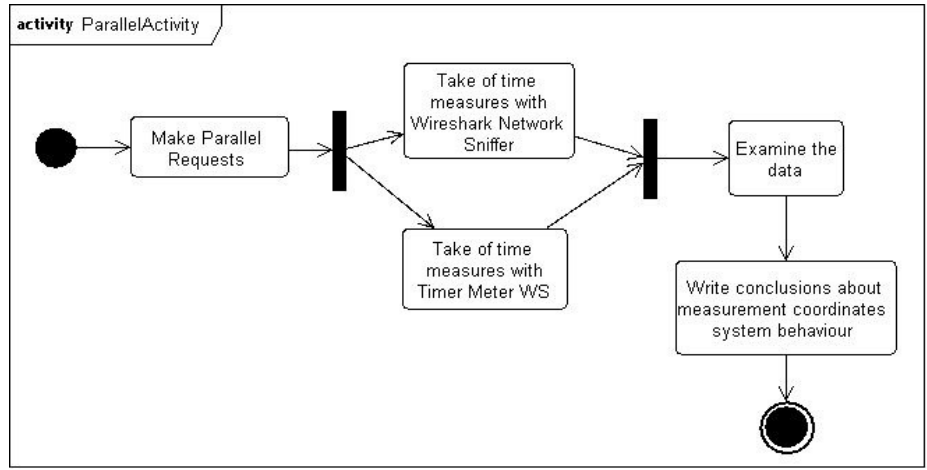

Figure 4. Parallel Requests. Activity Diagram

### 3.2 Localization Web Service

As discussed in the previous section, the coordinates calculated by the Vicon System can be sent through a web service. The Web Service provides the client its list of coordinates from the robot identifier. The lists of coordinates contains 7 values with the 3 robot positions (x,y,z), the angles for each axis and the yaw angle corresponding to the robot orientation respect the axis of coordinates. As can be seen in Figure 5, the web method named "getCoordinates" has the RobotID as an input parameter and returns a double list called "retorno" which contains the coordinates. The "LocalizationImplementation" class is connected to the Vicon System by a TCP socket in order to obtain the data.

```
@WebService()
public class LocalizationWS {

        @WebMethod(operationName = "getCoordinates")
        public double[] getCoordinates(@WebParam(name="RobotID")String RobotID) {
                LocalizationImplementation vc=new LocalizationImplementation();
                double[] retorno=vc.getCoordinates(RobotID);
                return retorno;
        }
}
```

Figure 5. Localization Web Service Invocation

### 3.3 Timer Meter Web Service

For the measuring time, a network sniffer is chosen, but the Localization Web Service requires a time for the method invocations which are not recorded by the sniffer. With the measurement obtained by the sniffer, a web service has been developed for calculating the time between the invocation of the method "getCoordinates" and the coordinates arrival to the client.

As can be seen in Figure 6, the Time Meter WS has three operations. The "startTime" method obtains the time at the invocation of the "getCoordinates" method into the Localization WS. The "finishTime" method obtains the time at the coordinates arrival to the robot. Finally the "time" method returns the time difference between the "getCoordinates" invocation and the coordinate's arrival.

International Journal of Distributed and Parallel Systems (IJDPS) Vol.2, No.2, March 2011

```
@WebService()
public class TimeMeter {

    @WebMethod(operationName = "startTime")
    public Long startTime() {
        long    start = System.currentTimeMillis();
        return start;
    }

    @WebMethod(operationName = "finishTime")
    public Long finishTime() {
        long end = System.currentTimeMillis();
        return end;
    }

    @WebMethod(operationName = "time")
    public Long time(@WebParam(name="f")long f, @WebParam(name="s")long s) {
        long t = (f-s);
        return t;
    }
}
```

Figure 6. Time Meter Web Service Implementation

## 4. RESULTS

This section shows the results obtained for the experiments described above. In the first section, are shown the results for the serial execution of coordinate's requests. Below are shown the result of parallel execution of coordinate's requests.

### 4.1 Serial executions

The first part of the experiment compares the response times consumed by the Web Service and the Socket. Table 2 shows the data obtained using the methodology explained in previous sections.

Table 2. Data collected by the Timer Meter WS

| Number of serial iterations | Average (ms) Socket | Average (ms) Web Services |
|---|---|---|
| 100 | 408 | 514 |
| 500 | 391 | 544 |
| 1000 | 391 | 564 |

These data confirm that the number of bytes required by Web Services is much larger than the required by Sockets (see Figure 7), which carries a higher response time (see Figure 8).

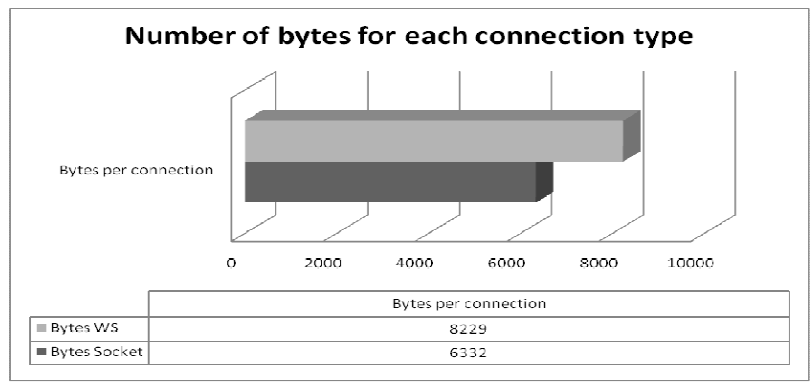

Figure 7. Number of bytes for each connection type

International Journal of Distributed and Parallel Systems (IJDPS) Vol.2, No.2, March 2011

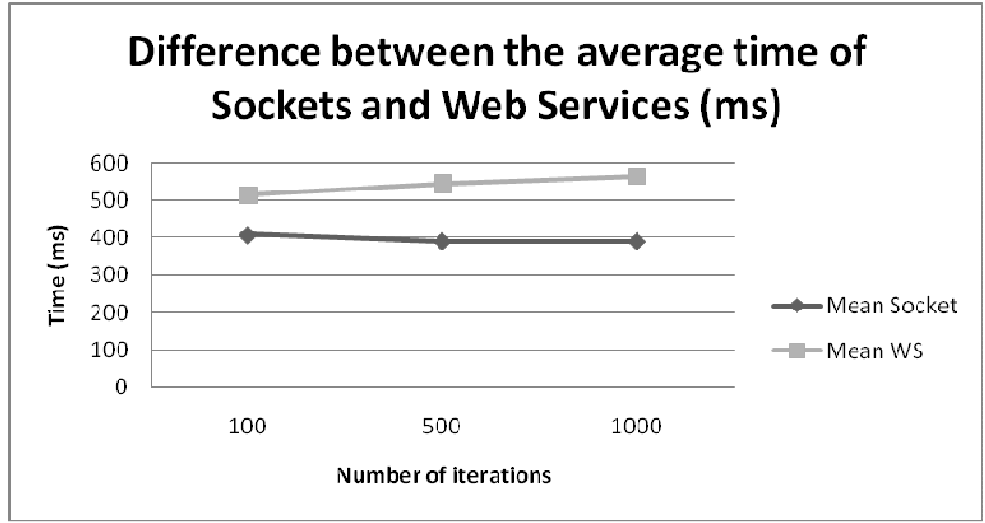

Figure 8. Average response time of Sockets & WS

The order of milliseconds difference between sockets and web services lead to the conclusion that the hypothesis stating that web services are not valid for real-time applications such as coordinate measuring is not true.

### 4.2 Parallel executions

In the second part of the experiment, each localization web service client runs different requests in parallel as described in the previous section.

Time difference between the averages calculated by the Timer Meter WS and the Wireshark (see Table 3) are due to the network sniffer which measures the time of the message request/response remains in the network. The Timer Meter WS starts the time measurement at the invocation of the method for obtaining the coordinates, and ends it when the web service client receives those coordinates.

Table 3. Data collected in parallel requests executions

| Number of serial iterations | Average (ms) Wireshark | Average (ms) Timer Meter WS |
|---|---|---|
| 5 | 374 | 443 |
| 10 | 390 | 453 |
| 20 | 470 | 522 |
| 50 | 405 | 450 |
| 100 | 469 | 509 |
| 250 | 476 | 511 |
| 500 | 491 | 524 |
| 1000 | 487 | 516 |

As shown in Figure 9, the average response time does not increase with the expected behaviour. Throughout the data collection, it has been observed that the coordinate measuring system (Vicon System) responds sequentially to parallel requests forming a chokepoint in the robot localization. Dependence that comes with the connection, ultimately, through a TCP socket explains the observed behaviour.

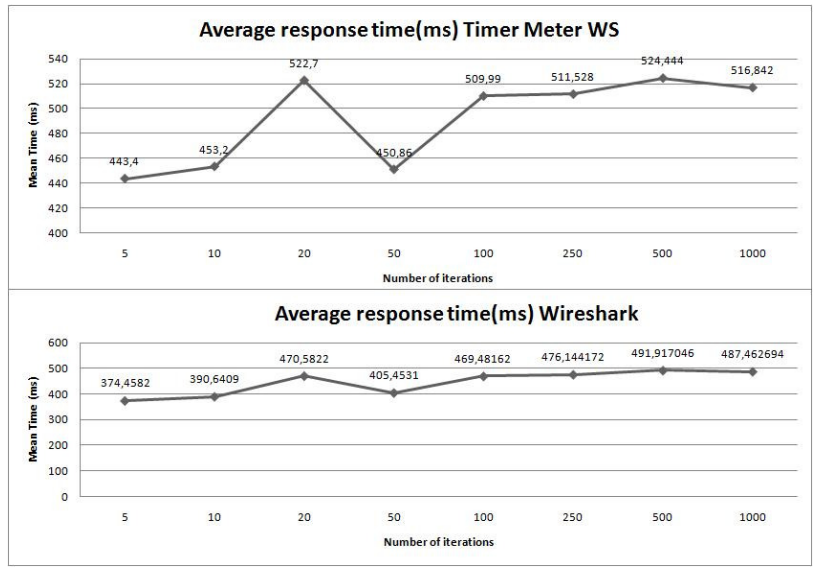

Figure 9. Average response time in parallel executions.

## 5. CONCLUSION AND FUTURE WORK

In the first part of the experiment can be concluded that web services are an optimal interoperability alternative, the time added respect to a socket connection is despicable for consider that web services add distort to the position measurement of a mobile device. The importance of this distortion is related with the robot speed.

The second part of the experiment shows the existent dependence between the robot localization system and the coordinate measurement adjacent system. The atomic access to the system that provides the coordinates, restricts the ability to respond to location requests executed in the same space of time with the assurance that the robot gets the approximate value of location at the time that the request was made. A coordinate measurement system based on active markers could solve the problem as well as any system capable of meeting requests in parallel.

In the future, it is necessary to solve problems of service requests in parallel. To solve this problem, a new coordinates measuring system can be designed based on other technologies such as OpenCV or RTLS, it is also possible to design an UDP socket connection to the Vicon System which would be able to solve broadcast coordinate requests. The Essex Robotics Research Group is working along with Infobotica Research Group on testing this new experiment performed on the Vicon System. A new article will be published about the results obtained on that work.

## ACKNOWLEDGEMENTS

Thanks to the financial support by the Spanish Department of Science and Technology. We must acknowledge the continuous support given by the University of Oviedo, and also for providing all the resources that made our research possible and, in particular, by their management of the DHCompliant project grant: MITC-09-TSI-020100-2009-359. Also appreciate the allowance of the Essex University and the Science Ministry of UK that made this research possible, and specially acknowledge the Human Centered Robotics (HCR) Group that lent their laboratory and their arena to begin with the experiments.

**Authors**

Alberto Alonso Fernández is a B.S. in ICT by the University of Oviedo. He holds the CCNA certification and has completed the Cisco official course of "Hardware and operating systems". He has worked in Inadeco S.L as a Microcomputer Systems Technician. Nowadays he's working in Infobotica Research Group, it's research group that belongs to the University of Oviedo, in the development of robotic and home automation software, within DHCompliant.

Omar Álvarez Fres obtained the title of Technician in Computer Applications Development in 2001, at Seresco College of Computing. He worked as a programmer in the Asturian Society of Industrial and Economic Studies (SADEI) and entered the University of Oviedo to study Computer Engineering Systems, obtaining the degree in 2009. He is member of Infobotica Research Group (linked to the University of Oviedo) and has participated in several research projects funded by the Spanish Ministry of Industry, Tourism and Trade. He has been grant holder in the Computer Services of University of Oviedo (Department of Statistics and Operations Research). He is also the author of "Rovim: A generic and extensible virtual machine for mobile robots", ICONS 2010. He also is studying a Master Degree in Web Engineering in the University of Oviedo (Research module) and he is also preparing his Final Master Thesis of Web Engineering techniques applied to Robotics. During the last years he has done additional University level courses on robotics, software development .NET, Semantic Web and R&D Project Management.

Dr. Ignacio González Alonso. He is an associated professor at the University of Oviedo since 2005. He also takes part in Infobotica Research Group at the University of Oviedo, developing projects in computer and robotics fields. He is collaborating in several national funded projects: TIC4Bot and DHCompliant, this year he is working on three proposals for the Seventh Framework Programme. He also is the co-author of twelve international contributions. He has a publication in the Inderscience publishers of distinguished academic, scientific and professional journals. He has previously worked in the private IT sector and he has managed his own company, what made him received the 2nd award for the best enterprise project.

Prof. Huosheng Hu. He leads the Human Centred Robotics (HCR) Group at Essex and has successfully completed a number of research projects funded by EU, EPSRC, The Royal Society and industry. His research interests include autonomous mobile robots, human-robot interaction, multi-robot collaboration, embedded systems, pervasive computing, sensor integration, evolutionary robotics, intelligent control and networked robotics. He received the MSc degree in industrial automation from the Central South University, China and the PhD degree in robotics from the University of Oxford. He has published over 300 papers in journals, books and conferences, and received a number of best paper awards. He is an Editor-in-Chief of Journal of Automation & Computing. He is a Chartered Engineer, a Fellow of IET, a senior member of IEEE and ACM.